\documentclass{article}

\usepackage{amsmath,amssymb}
\newtheorem{myDef}{Definition}
\newtheorem{myTheo}{Theorem}
\newtheorem{myProp}{Proposition}
\newtheorem{myCor}{Corollary}

\usepackage{times}
\usepackage{graphicx} 

\usepackage{pgfplots,pgfplotstable}

\pgfplotsset{
    jitter/.style={
        y filter/.code={\pgfmathparse{\pgfmathresult+#1}}
    },
    jitter/.default=0.1
}

\usepackage[numbers,sort&compress]{natbib}

\usepackage{algorithm}
\usepackage{algorithmic}

\usepackage{booktabs}
\usepackage{multirow}

\usepackage{rotating}

\DeclareMathOperator*{\argmax}{argmax}

\allowdisplaybreaks[3]

\title{Advances in\\ Learning Bayesian Networks\\of Bounded Treewidth}

\author{%
Siqi Nie\footnote{Email: nies@rpi.edu. Affiliation: Rensselaer
  Polytechnic Institute, USA.}\\
Denis Deratani Mau\'a\footnote{Email: denis.maua@usp.br. Affiliation:
  Universidade de S\~ao Paulo, Brazil.}\\
Cassio Polpo de Campos\footnote{Email: cassio@idsia.ch. Affiliation:  Dalle
  Molle Institute for Artificial Intelligence, Switzerland.}\\
Qiang Ji\footnote{Email: jiq@rpi.edu. Affiliation: Rensselaer Polytechnic Institute, USA.}
}

\date{\today}

\begin{document}

\maketitle

\begin{abstract} 
  This work presents novel algorithms for learning Bayesian network
  structures with bounded treewidth. Both exact and approximate methods
  are developed. The exact method combines mixed-integer linear
  programming formulations for structure learning and treewidth
  computation. The approximate method consists in uniformly sampling
  $k$-trees (maximal graphs of treewidth $k$), and subsequently
  selecting, exactly or approximately, the best structure whose moral
  graph is a subgraph of that $k$-tree. Some properties of these methods
  are discussed and proven. The approaches are empirically compared to
  each other and to a state-of-the-art method for learning bounded
  treewidth structures on a collection of public data sets with up to
  100 variables. The experiments show that our exact algorithm
  outperforms the state of the art, and that the approximate approach is
  fairly accurate.
\end{abstract} 

\section{Introduction}

Bayesian networks are graphical models widely used to represent joint
probability distributions on complex multivariate domains
\cite{koller2009probabilistic}. A Bayesian network comprises two parts:
a directed acyclic graph (the structure) describing the relationships
among variables in the model, and a collection of conditional
probability tables from which the joint distribution can be
reconstructed. As the number of variables in the model increases,
specifying the underlying structure becomes a tedious and difficult
task, and practitioners often resort to learning Bayesian networks
directly from data. Here, learning a Bayesian network refers to
inferring the underlying graphical structure from data, a task 
well-known to be NP-hard \cite{lh}.

Learned Bayesian networks are commonly used for drawing inferences such
as querying the posterior probability of some variable after evidence is
entered (a task known as belief updating), finding the mode of the joint
distribution (known as most probable explanation or MAP inference), or
selecting a configuration of a subset of the variables that maximizes
their conditional probability (known as marginal MAP inference). All
those inferences are NP-hard to compute even approximately
\cite{approx2,Roth96,approx1,darwiche2009,maphard}, and all known (exact
and provably good) algorithms have worst-case time complexity that is
exponential in the treewidth
\cite{koller2009probabilistic,darwiche2009,maua2012icml,ermon-et-al2013icml},
which is a measure of connectedness of the graph. Polynomial-time
algorithms for such inferences do exist, but they provide no guarantees
on the quality of the solution they deliver, which raises doubts as to
whether occasional bad results are a consequence of suboptimal structure
learning or of approximate inference. In fact, under widely believed
assumptions from complexity theory, exponential time complexity in the
treewidth is inevitable for any algorithm that provides provably good
inferences \cite{csh2008uai,kwisthout10}. Thus, learning network
structures of small treewidth is essential if one wishes to perform
reliable and efficient inference. This is particularly important in the
presence of missing data, as learning methods usually resort to some
kind of Expectation-Maximization procedure that requires performing
belief updating in the network at every iteration
\cite{friedman1998uai}. In those cases inefficient inference leads to
great computational cost of learning; unreliable inference leads to learning
underfitted/overfitted structures.

Since estimating a network's treewidth is itself an NP-hard task
\cite{twhard}, extending current methods for learning Bayesian networks
to the case of bounded treewidth while maintaining their relative
efficiency and accuracy is not trivial. In comparison to unconstrained
Bayesian network learning, few algorithms have been designed for the
bounded treewidth case. \citet{korhonen2exact} showed that learning
bounded treewidth Bayesian networks is NP-hard, and developed an exact
algorithm based on dynamic programming that learns optimal $n$-node
structures of treewidth at most $\omega$ in time $3^n n^{\omega+O(1)}$,
which is above the $2^n n^{O(1)}$ time required by the best worst-case
algorithms for learning optimal Bayesian networks with no constraint on
treewidth \cite{silander-myllymaki2006uai}. \citet{elidan} combined
several heuristics to treewidth computation and network structure
learning in order to design approximate methods. Others have addressed
the similar (but not equivalent) problem of learning undirected models
of bounded treewidth
\cite{bach2001thin,srebro2003maximum,chechetka2007efficient}. Very
recently, there seems to be an increase of interest in the
topic. \citet{bjm2014aistats} showed that the problem of learning
bounded treewidth Bayesian networks can be reduced to a weighted maximum
satisfiability problem, and subsequently solved by weighted MAX-SAT
solvers. They report experimental results showing that their approach
outperforms \citeauthor{korhonen2exact}'s dynamic programming
approach. In the same year, \citet{pfl2014aistats} showed that the
problem can be reduced to a mixed-integer linear program (MILP), and
then solved by off-the-shelf MILP optimizers (e.g.~CPLEX). Their reduced
MILP problem however has exponentially many constraints in the number of
variables. Following the work of \citet{cussens2}, the authors avoid
creating such large programs by a cutting plane generation mechanism,
which iteratively includes a new constraint while the optimum is not
found. The generation of each new constraint (cutting plane) requires
solving another MILP problem. The works of \cite{bjm2014aistats} and
\cite{pfl2014aistats} have been developed independently and
simultaneously with our work presented here; for this reason, we do not
compare our methods with theirs. We intend to do so in the near future.

In this paper, we present two novel ideas for score-based Bayesian
network structure learning with a hard constraint on treewidth.  We
first introduce a mixed integer linear programming formulation of the
problem (Section~\ref{sec3}) that builds on existing MILP formulations
for unconstrained structure learning of Bayesian
networks~\cite{cussens2,cussens1} and for computing the treewidth of a
graph~\cite{grigoriev2011integer}. The designed formulation is able to
find a score-maximizer Bayesian network of treewidth smaller than a
given constant for models containing many more variables
than~\citeauthor{korhonen2exact}'s method, as we empirically demonstrate
in Section~\ref{sec5}. Unlike the MILP formulation of
\citet{pfl2014aistats}, the MILP problem we generate is of polynomial
size in the number of variables, and does not require the use of cutting
planes techniques. This makes for a clean and succinct formulation that
can be solved with a single call of a MILP optimizer. A better
understanding of cases where one approach is preferred to the other is
yet to be achieved.

Since linear programming relaxations are used for solving the MILP
problem, any MILP formulation can be used to provide approximate
solutions and error estimates in an anytime fashion (i.e., the method
can be stopped at any time during the computation with a feasible
solution). However, the MILP formulations (both ours and the one
proposed by \citet{pfl2014aistats}) cannot cope with very large domains,
even if we agreed on obtaining only approximate solutions. This is
because the minimum size of the MILP problems is cubic in the number of
variables (hence it is difficult even to start the MILP solver for large
domains), and there is probably little we can do to considerably improve
this situation (a further discussion on that is given in
Section~\ref{sec3}). This limitation is observed in the experiments
reported in Section~\ref{sec5}, where our MILP formulation requires a
much larger amount of time to obtain much poorer solutions for networks
with over 50 variables.

In order to deal with large domains, we devise (in Section~\ref{sec4})
an approximate method based on a uniform sampling of $k$-trees (maximal
triangulated graphs of tree\-width $k$), which is achieved by using a
fast computable bijection between $k$-trees and Dandelion
codes~\cite{caminiti2010bijective}. For each sampled $k$-tree, we either
run an exact algorithm similar to the one proposed
in~\cite{korhonen2exact} (when computationally appealing) to learn the
score-maximizing network whose moral graph is a subgraph of that
$k$-tree, or we resort to a much more efficient method that takes
partial variable orderings uniformly at random from a (relatively small)
space of orderings that are compatible with the $k$-tree. We discuss the
time and sample complexity of both variants, and compare it to those of
similar schemes for learning unconstrained networks. We show empirically
(in Section~\ref{sec5}) that the double sampling scheme (of $k$-trees
and partial variable orderings) is very effective in learning close to
optimal structures in a selected set of data sets. We conclude in
Section~\ref{sec6} by noting that the methods we propose can be
considered as state-of-the-art, and by suggesting possible improvements.
To start, Section~\ref{sec2} presents some background knowledge on
learning Bayesian networks.

\section{Preliminaries}
\label{sec2}

A Bayesian network is a concise graphical representation of a
multivariate domain, where each random variable is associated with a
node of its underlying directed acyclic graph (DAG) and local
conditional probability distributions are specified for the variable
given its parents in the graph (we often refer to variables and nodes in
the graph interchangeably). 

Let $N$ be $\{1,\ldots,n\}$ and consider a finite set $X=\{X_i:i\in N\}$
of categorical random variables $X_i$ taking values in finite sets
$\mathcal{X}_i$. Formally, a Bayesian network is a triple
$(X,G,\theta)$, where $G=\{N,A\}$ is a DAG whose nodes are in one-to-one
correspondence with variables in $X$, and $\theta=\{\theta_i(x_i,
x_{\pi_i})\}$ is a set of numerical parameters specifying (conditional)
probability values $\theta_i(x_i,x_{\pi_i})=P(x_i|x_{\pi_i})$, for every
node $i$ in $G$, value $x_i$ of $X_i$ and assignment $x_{\pi_i}$ to the
parents $\pi_i$ of $X_i$, according to $G$. The structure $G$ (that is,
the DAG of the network) represents a set of stochastic independence
assessments among variables in $X$. In particular, $G$ represents a set
of graphical Markov conditions: every variable $X_i$ is conditionally
independent of its nondescendant nonparents given its parents. As a
consequence, a Bayesian network $(X, G, \theta)$ uniquely defines a
joint probability distribution over $X$ as
the product of its parameters~\cite[Chapter 3.2.3]{koller2009probabilistic}:
\begin{equation}
\label{eqn: joint}
P(x_1,\ldots,x_n; G,\theta)=\prod_{i\in N}\theta_i(x_i, x_{\pi_i}).
\end{equation}

Learning the structure $G$ from data is a challenging problem. One
approach is to identify, for each variable, the minimal set of variables
that makes that variable conditionally independent of others (Markov
blanket), which is usually done by means of statistical tests of
stochastic independence or information theoretic
measures~\cite{spirtes1995learning}. Alternatively, structural learning
can be posed as a combinatorial optimization problem in which one seeks
the structure that maximizes a score function that relates to the data
likelihood, while avoiding some excessive model complexity. Commonly
used score functions include the Minimum Description Length (which is
equivalent to the Bayesian Information Criterion)
\cite{schwarz1978estimating}, and Bayesian Dirichlet (likelihood)
equivalent uniform score
\cite{buntine1991theory,cooper1992bayesian,heckerman1995learning}. These
functions follow different rationale but they all satisfy two
properties: (i) they can be written as a sum of local score functions
that depend only on the parent set of each node and on the data, and
(ii) the local score functions can be efficiently computed and stored.
Score-based structure learning is a difficult task, and research on this
topic has been very active
\cite{jaakkola-et-al2010aistats,Hemmecke20121336,Yuan12improved,cussens1,cussens3,Yuan13learning,korhonen2exact}.

In score-based Bayesian network learning we seek a DAG structure $G^*$
such that
\begin{equation}
G^*=\argmax_{G\in \mathcal{G}_n} \sum_{i\in N} s_i(\pi_i) \, , 
\end{equation}
where $\mathcal{G}_n$ is the class of all DAGs with $n$ nodes, $s_i$ are
local score functions that depend only on the parent set $\pi_i$ as
given by $G$ (i.e., the computation of each $s_i(\pi_i)$ depends only on
the values that $X_i$ and $X_{\pi_i}$ take in the data set). We assume
(unless otherwise stated) that local scores $s_i(\pi_i)$ have been
previously computed and can be retrieved at constant time. Despite the
decomposability of the score functions, the optimization cannot be
performed locally lest it almost certainly introduce directed cycles in
the graph. 

We say that a cycle in an undirected graph has a chord if there are two
nodes in the cycle which are connected by an edge outside the cycle. A
chordal graph is an undirected graph in which all cycles of length four
or more have a chord. Any graph can be made chordal by inserting edges,
a process called \emph{chordalization} \cite{twhard,bodtw}. The treewidth of a
chordal graph is the size of its largest clique minus one. The treewidth
of an arbitrary undirected graph is the minimum treewidth over all
chordalizations of it. The moral graph of a DAG is the undirected graph
obtained by connecting any two nodes with a common child and dropping
arc directions. The treewidth of a DAG is the treewidth of its
corresponding moral graph. The treewidth of a Bayesian network
$(X,G,\theta)$ is the treewidth of the DAG $G$.

An \emph{elimination order} is a linear ordering of the nodes in a
graph. We say that an elimination order is \emph{perfect} if for every
node in the order its higher-ordered neighbors form a clique (i.e., are
pairwise connected). A graph admits a perfect elimination order if and
only if it is chordal. Perfect elimination orders can be computed in
linear time if they exist. The \emph{elimination} of a node according to
an elimination order is the process of pairwise connecting all of its
higher-ordered neighbors. Thus, the elimination of all nodes produces a
chordal graph for which the elimination order used is perfect. The edges
inserted by the elimination process are called \emph{fill-in}
edges. Given a perfect elimination order, the treewidth of the graph can
be computed as the maximum number of higher ordered neighbors in the
graph. 

The reason why most score functions penalize model complexity (as given
by the number of free numerical parameters) is that data likelihood
always increases by augmenting the number of parents of a variable (and
hence the number of free parameters in the model), which leads to
overfitting and poor generalization. The way scores penalize model complexity
generally leads to structures of bounded in-degree and helps in
preventing overfitting, but even bounded in-degree graphs can have large
treewidth (for instance, directed square grids have treewidth equal to
the square root of the number of nodes, yet have maximum in-degree equal
to two), which yields a great problem to subsequent probabilistic
inferences with the model.

There are at least two direct reasons to aim at learning Bayesian
networks of bounded treewidth: (i) As discussed previously, all known
exact algorithms for probabilistic inference have exponential time
complexity in the treewidth, and networks with very high treewidth are
usually the most challenging for approximate methods; (ii) Previous
empirical results~\cite{perrier2008finding,elidan} suggest that bounding
the treewidth might improve model performance on held-out data. There is
also evidence that bounding the treewidth does not impose a great burden
on the expressivity of the model for real data
sets~\cite{br2003nips}. 

The goal of learning Bayesian networks of bounded treewidth is to search
for $G^*$ such that
\begin{equation} \label{obj-btw}
G^*=\argmax_{G\in \mathcal{G}_{n,k}} \sum_{i\in N} s_i(\pi_i) \, ,
\end{equation}
where $\mathcal{G}_{n,k}$ is the class of all DAGs of treewidth not
(strictly) greater than $k$. From a theoretical point of view, this is
no easy task. \citet{korhonen2exact} adapted
\citeauthor{srebro2003maximum}'s complexity result for Markov networks
\cite{srebro2003maximum} to show that learning the structure of Bayesian
networks of bounded treewidth strictly greater than one is NP-hard.
\citeauthor{dasgupta}'s results also prove this hardness if the score
maximizes data likelihood~\cite{dasgupta} (in the case of networks of
treewidth one, that is, directed trees with at most one parent per node,
learning can be performed efficiently by the Chow and Liu's algorithm
\cite{chow1968approximating}).

\section{Mixed integer linear programming}
\label{sec3}

The first contribution of this work is the mixed integer linear
programming (MILP) formulation that we design to exactly solve the
problem of structure learning with bounded treewidth. MILP formulations
have shown to be very effective to learning Bayesian networks without
the treewidth bound \cite{cussens2,cussens3}, surpassing other attempts
in a range of data sets. Moreover, the great language power of a MILP
problem allows us to encode the treewidth constraint in a natural
manner, which might not be easy with other structure learning
approaches~\cite{Yuan13learning,Yuan12improved,Hemmecke20121336,decampos09icml,koivisto09uai}. We
note that computing the treewidth of a graph is an NP-hard problem
itself \cite{twhard}, even if there are linear algorithms that are
\emph{only} exponential in the treewidth \cite{bodtw} (these algorithms
might be seen mostly as theoretical results, since their practical use
is shadowed by very large hidden constants). Hence, one should not hope
to enforce a bound on the treewidth (which should work for any chosen
bound) without a machinery that is not at least as powerful as NP.

The novel formulation is based on combining the MILP formulation for
structure learning in \cite{cussens1} with the MILP
formulation presented in~\cite{grigoriev2011integer} for computing the
treewidth of an undirected graph. There are although crucial
differences, which we highlight later on. We have avoided the use of
sophisticated techniques for MILP in the context of structure learning,
such as constraint generation~\cite{cussens2,cussens3}, because we are
interested in providing a clean and succinct MILP formulation, which can be
ran using off-the-shelf solvers without additional coding.

Since our formulation is a combination of two previous MILP formulations of
distinct problems, we will present each formulation separately, and
then describe how to combine them into a concise MILP problem. 

\subsection{A MILP formulation for bounding the treewidth}

Consider a graph $G=(N,E)$. We
begin with the MILP formulation of the class of all supergraphs of a
graph $G$ that have treewidth less than or equal to a given value $w$:
\begin{subequations} \label{conmilp1}
\begin{align}
\label{con1}\sum_{j\in N}y_{ij} &\leq w,&&\forall i\in N,\\
\label{con4}(n+1)\cdot y_{ij}&\leq n + z_{j}-z_{i}, &&
\forall i,j \in N,\\
\label{con5}y_{ij}+y_{ji} &= 1,&& \forall (i,j) \in E,\\
\label{con6}y_{ij}+y_{ik}-(y_{jk}+y_{kj})&\leq 1,&& \forall i,j,k\in N,\\
\label{con7}z_{i}\in [0,n], &&& \forall i \in N,\\
\label{con8} y_{ij}\in \{0,1\},&&&\forall i,j \in N.
\end{align}
\end{subequations}
The formulation above is based on encoding all possible elimination
orders of the nodes of $G$. A chordalization $G'=(N,E')$ of $G$ of
treewidth at most $w$ can be obtained from a feasible solution (if it
exists) of the program by setting
$E'=\{ij \in N \times N: y_{ij} = 1 \text{ or } y_{ji}=1\}$. Constraint
\eqref{con1} ensures $G'$ has treewidth at most $w$ by bounding the
number of higher-ordered neighbors of every node $i$ (which is an
alternative way of defining the treewidth of chordal graphs).
The variables $z_i$, $i \in N$, take (real) values in $[0,n]$
(Constraint \eqref{con7}) and partially define an elimination order of
the nodes: a node $i$ is eliminated before node $j$ if $z_i < z_j$ (the
specification is partial since its allows for two nodes $i$ and $j$ with
$z_i=z_j$). This order does not need to be linear because there are
cases where multiple linearizations of the partial order are equally
good in building a chordalization $G'$ of $G$ (i.e., in minimizing the
maximum clique size of $G'$). In such cases, two nodes $i$ and $j$ might
be assigned the same value $z_i=z_j$ indicating that eliminating $z_i$
before $z_j$ and the converse results in chordal graphs with the same
treewidth.  The variables $y_{ij}$, $i,j \in N$, are $\{0,1\}$-valued
(Constraint \eqref{con8}) and indicate whether node $i$ precedes $j$ in
the order (i.e., whether $z_i < z_j$) \emph{and} an edge exists among
them in the resulting chordal graph $G'$ (recall that an elimination
process always produces a chordal graph).
Although the values $z_i$ are not forced to be integers in our
formulation, in practice they will most likely be so. Constraint \eqref{con4}
allows $y_{ij}$ to be $1$ only if $j$ appears after $i$ in the order (it
in fact requires that $z_j\geq z_i+1$ to allow $y_{ij}$ to be
one). Constraint \eqref{con5} ensures $G'$ is a supergraph of
$G$. Constraint \eqref{con6} guarantees that the elimination ordering
induced by $z_i$, $i\in N$, is perfect for $G'$: if $j$ and $k$ are
higher ordered neighbors of $i$ in $G'$, then $j$ and $k$ are also
neighbors in $G'$, that is, either $y_{jk}$ or $y_{kj}$ must be $1$. The
practical difference of this formulation with respect to the one
in~\cite{grigoriev2011integer} lies in the fact that we allow partial
elimination orders, and we do not need integer variables to enforce such
orders. A bottleneck is the specification of Constraint~\eqref{con6}, as
there are $\Theta(n^3)$ such constraints. The following result is an
immediate conclusion of the above reasoning.

\begin{myProp} \label{MILP1-is-correct} The graph $G$ has treewidth at
  most $w$ if and only if the set defined by Constraints
  \eqref{conmilp1} is non empty.
\end{myProp}

\begin{myProp} \label{MILP1-outputs-chordalization} Let $z_i, y_{ij}$, $i,j \in N$,
  be variables satisfying Constraints \eqref{con1}--\eqref{con8}. Then
  the graph $G'=(E',N)$, where
  $E'=\{ij \in N \times N: y_{ij} = 1 \text{ or } y_{ji}=1\}$, is a
  chordalization of $G$ with treewidth at most $w$, and any elimination
  order consistent with the partial order induced by $z_i$ is 
  perfect for $G'$.
\end{myProp}

\subsection{A MILP formulation for structure learning}

We now turn our attention to the MILP formulation of the structure
learning part. Consider a chordal (undirected) graph $M=(N,E)$, a
perfect elimination order for $M$, and let $y_{ij}$, $i,j \in N$, be
$\{0,1\}$-valued variables such that $y_{ij}=1$ if and only if $E$
contains $ij$ and $i$ is eliminated before $j$. For each node $i$ in $N$
let $\mathcal{F}_i$ be the collection of all allowed parent sets for
that node (these sets can be specified manually by the user or simply
defined as the subsets of $N \setminus \{i\}$ with cardinality less than
a given bound). We denote an element of $\mathcal{F}_i$ as $F_{it}$,
with $t=1,\dotsc,|\mathcal{F}_i|$ (hence $F_{it} \subset N$). The
following MILP formulation specifies the class of all DAGs over $N$ that
are consistent with the parent sets $\mathcal{F}_i$ and whose moral
graph is a subgraph of $M$:
\begin{subequations} \label{conmilp2}
\begin{align}
\label{con10}\sum_t \pi_{it} &= 1, && \forall i\in N,\\
\label{con11}(n+1)\pi_{it} &\leq n +
v_{j}-v_{i}, &&\forall i
\in N,\,\forall t,\,\forall j \in F_{it},\\
\label{con12}\pi_{it}&\leq y_{ij}+y_{ji}, && \forall i
\in N,\,\forall t,\,\forall j \in F_{it},\\
\label{con13}\pi_{it}&\leq y_{jk}+y_{kj}, && \forall i
\in N,\,\forall t,\,\forall j,k \in F_{it},\\
\label{con14}v_{i}\in [0,n], &&& \forall i \in N,\\
\label{con15} \pi_{it}\in \{0,1\}, &&&
\forall i\in N,\, \forall t,
\end{align}
\end{subequations}
\noindent where the scope of the $\forall t$ in each constraint is
$1,\ldots,|\mathcal{F}_i|$. A DAG $D=(N,A)$ can be obtained from a
solution to the above program by setting
$A=\{ i \leftarrow j: i \in N, j \in N, \pi_{it}=1 \text{ and
} j \in F_{it} \}$. The variables $v_i$, $i \in N$, take values in
$[0,n]$ (Constraint \eqref{con14}) and partially specify a topological
order of the nodes in $D$: if $v_i > v_j$ then $j$ is not an ancestor of
$i$. The variables $\pi_{it}$, $i\in N$, $t=1,\dotsc,|\mathcal{F}_i|$,
are $\{0,1\}$-valued (Constraint \eqref{con15}) and indicate whether the
$t$-th parent set in $\mathcal{F}_i$ was chosen for node $i$. Constraint
\eqref{con10} enforces that exactly one parent set is chosen for each
node. Constraint \eqref{con11} forces those choices to be acyclic, that
is, to respect the topological order induced by the variables $v_i$
(with ties broken arbitrarily for nodes $i,j$ with $v_i = v_j$). Here
too the order does not need to be linear. In fact, only the relative
ordering of nodes that are connected in $M$ is relevant because
Constraints \eqref{con12} and \eqref{con13} ensure that arcs appear in
$D$ only if the corresponding edges in the moral graph of $D$ exist in
$M$ (Constraint \eqref{con13} is responsible for having the moralization
of the graph falling inside $M$).

\begin{myProp} \label{MILP2-is-correct} Let $v_i$, $\pi_{it}$,
  $i \in N, t=1,\dotsc,|\mathcal{F}_i|$, be variables satisfying
  Constraints \eqref{conmilp2}. Then the directed graph $D=(N,A)$,
  where $A=\{ i \leftarrow j: i \in N, j \in N, \pi_{it}=1 \text{ and
  } j \in F_{it} \}$ is acyclic and consistent with every set
  $\mathcal{F}_i$. Moreover the moral graph of $D$ is a subgraph of $M$.
\end{myProp}

A corollary of the above result is that the treewidth of $D$ is at most
the treewidth of $M$ \cite{bodtw}.

\subsection{Combining the MILP formulations}

We can now put together the two previous MILP formulations to reach the
following MILP formulation for the problem of learning DAGs of 
treewidth bounded by a constant $w$:

\begin{subequations} \label{conmilp}
\begin{align}
\intertext{maximize:}
\label{obj} & \sum_{it} \pi_{it}\cdot s_i(F_{it})\\
\intertext{subject to:}
\label{con1b}\sum_{j\in N}y_{ij} &\leq w,&&\forall i\in N,\\
\label{con4b}(n+1)\cdot y_{ij}&\leq n + z_{j}-z_{i}, &&
\forall i,j \in N,\\
\label{con6b}y_{ij}+y_{ik}-(y_{jk}+y_{kj})&\leq 1,&& \forall i,j,k\in N,\\
\label{con10b}\sum_t \pi_{it} &= 1, && \forall i\in N,\\
\label{con11b}(n+1)\pi_{it} &\leq n +
v_{j}-v_{i}, &&\forall i
\in N,\,\forall t,\,\forall j \in F_{it},\\
\label{con12b}\pi_{it}&\leq y_{ij}+y_{ji}, && \forall i
\in N,\,\forall t,\,\forall j \in F_{it},\\
\label{con13b}\pi_{it}&\leq y_{jk}+y_{kj}, && \forall i
\in N,\,\forall t,\,\forall j,k \in F_{it},\\
\label{con7b}z_{i}\in [0,n], & v_{i}\in [0,n], && \forall i \in N,\\
\label{con8b} y_{ij}\in \{0,1\},&&&\forall i,j \in N,\\
\label{con15b} \pi_{it}\in \{0,1\}, &&&
\forall i\in N,\, \forall t.
\end{align}
\end{subequations}

As the following result shows, the MILP formulation above specifies DAGs
of bounded treewidth: 

\begin{myTheo} \label{MILP-is-correct} Let $y_{ij},z_i,v_i,\pi_{it}$,
  $i,j \in N, t=1,\dotsc,|\mathcal{P}_i|$, be variables satisfying
  Constraints \eqref{con1b}--\eqref{con15b}, and define a directed graph
  $D=(N,A)$, where
  $A=\{ i \leftarrow j: i \in N, j \in N, \exists t \text{ s.t. } \pi_{it}=1 \text{ and
  } j \in F_{it} \}$. Then $D$ is a acyclic, consistent with the parents
  sets $\mathcal{P}_i$, and has treewidth at most $w$.
\end{myTheo}

\begin{myCor} \label{MILP-finds-optimum} If $y_{ij},z_i,v_i,\pi_{it}$,
  $i,j \in N, t=1,\dotsc,|\mathcal{P}_i|$, maximize \eqref{obj} and
  satisfy \eqref{con1b}--\eqref{con15b}, then the DAG $D$ as defined
  above is the solution to the optimization in \eqref{obj-btw}.
\end{myCor}

The MILP formulation \eqref{conmilp} can be directly fed into any
off-the-shelf MILP optimizer. According to Corollary
\eqref{MILP-finds-optimum}, the outcome will always be an optimum
structure if enough resources (memory and time) are given. Standard MILP
optimizers (e.g.~CPLEX) often employ branch-and-bound (or
branch-and-cut) procedures, which are able to be halted prematurely at
any time and still provide a valid solution and an outer bound for the
maximum score. Hence, the MILP formulation also provides an anytime
algorithm for learning Bayesian networks of bounded treewidth: the
procedure can be stopped at time and still provide an approximate
solutions and error bound. Moreover, the quality of the approximation
solution returned increases with time, while the error bounds
monotonically decrease and eventually converge to zero.

\subsection{Comparison with the dynamic programming approach}

To validate the practical feasibility of our MILP formulation, we
compare it against the the dynamic programming method proposed
previously for this problem~\cite{korhonen2exact}, which we call K\&P
from now on.\footnote{We used the freely available code provided by the
  authors at http://www.cs.helsinki.fi/u/jazkorho/aistats-2013/.}
Table~\ref{table1} show the time performance of our MILP formulation and
that of K\&P on a collection of reasonably small data sets from the UCI
repository\footnote{Obtained from http://archive.ics.uci.edu/ml/.}
(discretized over the median value, when needed) and small values of the
treewidth bound. More details about these data are presented in
Section~\ref{sec5}.  The experiments have been run with a limit of 64GB
in memory usage and maximum number of parents per node equal to three
(the latter restriction facilitates the experiments and does not impose
a constraint in the possible treewidths that can be found). While one
shall be careful when directly comparing the times between methods, as
the implementations use different languages (we are running CPLEX 12.4,
K\&P uses a Cython\footnote{http://www.cython.org.} compiled Python
code), we note that our MILP formulation is orders of magnitude faster
than K\&P, and able to solve many problems which the latter could not
(in Section~\ref{sec5} we show the results of experiments with much
larger domains). A time limit of 3h was given to the MILP, in which case
its own estimation of the error is reported (in fact, it found the
optimal structure in all instances, but was not able to certify it to
be optimal within 3h).

\begin{table}[ht]
\centering
\caption{Computational time to find the optimal Bayesian network
  structure. Empty cells indicate that the method failed to solve the instance
  because of excessive memory consumption. Limit of 3h was given
  to MILP, in which case its own estimation of the error is
  reported. $s,m,h$ mean seconds, minutes and hours, respectively.}
\begin{tabular}{ccllll}
\toprule
METHOD & TW & NURSERY & BREAST & HOUSING & ADULT \\
& & $n\!\!=\!\!9$ & $n\!\!=\!\!10$ & $n\!\!=\!\!14$ & $n\!\!=\!\!15$\\
\midrule
\multirow{4}{1.5cm}{MILP}& 2 & 1s & 31s &  3h [2.4\%] &  3h [0.39\%] \\
& 3 & $<$1s & 19s & 25m &3h [0.04\%] \\
& 4 &$<$1s & 8s & 80s & 40m\\
& 5 &$<$1s & 8s & 56s & 37s\\
\midrule 
\multirow{4}{1.5cm}{K\&P}& 2 & 7s & 26s & 128m & 137m \\
& 3 & 72s & 5m &--& --\\
& 4 & 12m & 103m &--&--\\
& 5 & 131m & -- & -- &--\\
\bottomrule
\end{tabular}
\label{table1}
\end{table}

The results in the table show that our MILP formulations largely
outperforms K\&P, being able to handle much larger problems. Yet we see
from these experiments that both algorithms scale poorly in the number
of variables. In particular, K\&P cannot cope with data sets containing
more than a dozen of variables. The results suggest that the MILP
problems become easier as the treewidth bound increases. This is likely
a consequence of the increase of the space of feasible solutions, which
makes the linear relaxations used for solving the MILP problem tighter,
thus reducing the computational load. This is probably aggravated by the
small number of variables in these data sets (hence, by increasing the
treewidth we effectively approximate an unbounded learning situation).

We shall demonstrate empirically in Section~\ref{sec5} that the quality
of solutions found by the MILP approach in a reasonable amount of
time degrades quickly as the number of variables reaches several
dozens. Indeed, the MILP formulation is unable to find reasonable
solutions for data sets containing 100 variables, which is not
surprising given that number of Constraints \eqref{con6b} and
\eqref{con13b} is cubic in the number of variables; thus, as $n$
increases even the linear relaxations of the MILP problem become hard to
solve. In the next section, we present a clever sampling algorithm over
the space of $k$-trees to overcome such limitations and handle large
domains. The MILP formulation just described will set a baseline for the
performance of such approximate approach.

\section{Sampling $k$-trees using Dandelion codes}
\label{sec4}

In this section we develop an approximate method for learning bounded
treewidth Bayesian networks that is based on sampling graphs of bounded
treewidth and subsequently finding DAGs whose moral graph is a subgraph
of that graph. The approach is designed aiming at data sets with large
domains, which cannot be handled by the MILP formulation.

A naive approach to designing an approximate method would be to extend
one of the sampling methods for unconstrained Bayesian network
learning. For instance, we could envision a rejection sampling approach,
which would sample structures using some available procedure (for
instance, by sampling topological orderings and then greedily finding a
DAG structure consistent with that order, as in
\cite{teyssier2012ordering}), and verify their treewidth, discarding the
structure when the test fails. There are two great issues with this
approach: (i) the computation of treewidth is a hard problem, and even
if there are linear-time algorithms (but exponential on the treewidth),
they perform poorly in practice; (ii) virtually all structures would be
discarded due to the fact that complex structures tend to have larger
scores than simple ones, at least for the most used score functions
(their penalizations reduce the local complexity of the model, but are
not able to constrain a global property such as treewidth). We
empirically verified these facts, but will not report further on them
here.

Another natural approach to the problem is to consider both an
elimination order for the variables (from which the treewidth can be
computed) and a topological order (from which one can greedily search
for parent sets without creating cycles in the graph). It is
straightforward to uniformly sample from the space of orderings, but the combined overall number
of such orderings is quite high: $(n!)^2\approx e^{2n\log n - 2n}$ (from
the Stirling approximation). We propose an interesting way that is more
efficient in terms of the size of the sampling space, and yet can be
sampled uniformly (uniform
sampling is a desirable property, as it ensures a good coverage of the
space and is superior to other options if one has no prior information
about the search space).  This approach is based on the set of $k$-trees.
\begin{myDef} A $k$-tree is defined in the following recursive way:\\
(1) A $(k+1)$-clique is a $k$-tree.\\
(2) If $T_k^{\prime}=(V,E)$ is a $k$-tree with nodes $V$ and edges $E$, $K\subseteq V$ is a $k$-clique
and $v\in N\setminus V$, then $T_k=(V\cup \{v\},E\cup \{(v,x)|x\in
K\})$ is a $k$-tree.
\label{def1}
\end{myDef}
We denote by $\mathcal{T}_{n,k}$ the set of all $k$-trees over $n$
nodes. In fact, a Bayesian network with treewidth bounded by $k$ is
closely related to a $k$-tree. Because $k$-trees are exactly the
maximal graphs with treewidth $k$ (graphs to which no more edges can
be added without increasing their treewidth), we know that the moral
graph of the optimal structure has to be a subgraph of a
$k$-tree~\cite{korhonen2exact}.

The idea is to sample $k$-trees and then search for the best structure
whose moral graph is one of the subgraphs of the $k$-tree.  While
directly sampling a $k$-tree might not be trivial,
\citet{caminiti2010bijective} proposed a linear time method for coding
and decoding $k$-trees into what is called {\it Dandelion} codes (the
set of such codes is denoted by $\mathcal{A}_{n,k}$). Moreover, they
established a bijective mapping between codes in $\mathcal{A}_{n,k}$ and
$k$-trees in $\mathcal{T}_{n,k}$.  The code $(Q,S)\in\mathcal{A}_{n,k}$
is a pair where $Q\subseteq N$ with $|Q|=k$ and $S$ is a list of $n-k-2$
pairs of integers drawn from $N\cup\{\epsilon\}$, where $\epsilon$ is an
arbitrary number not in $N$. For example, $Q=\{2,3,9\}$ and
$S=[(0, \epsilon), (2,1), (8,3), (8,2), (1,3), (5,3)]$ is a Dandelion
code of a (single) $3$-tree over $11$ nodes (that is , $n=11$,
$k=3$). Dandelion codes can be sampled uniformly at random by a trivial
linear-time algorithm that uniformly chooses $k$ elements out of $N$ to
build $Q$, and then uniformly samples $n-k-2$ pairs of integers in
$N\cup\{\epsilon\}$.

\begin{myTheo}
\cite{caminiti2010bijective} There is a bijection mapping elements of
$\mathcal{A}_{n,k}$ and $\mathcal{T}_{n,k}$ that is computable in time
linear in $n$ and $k$.\label{th0}
\end{myTheo}

Given $T_k\in\mathcal{T}_{n,k}$, we can use the dynamic programming
algorithm proposed in~\cite{korhonen2exact} to find the optimal
structure whose moral graph is a subgraph of $T_k$. Our implementation
follows the ideas in~\cite{korhonen2exact}, but can also be seen as
extending the divide-and-conquer method of~\cite{citeulike:100302} to
account for all possible divisions of nodes. This results in the
following theorem.
\begin{myTheo}
  \cite{korhonen2exact} For any fixed $k$, given (a $k$-tree)
  $G=(N, E)\in\mathcal{G}_{n,k}$ and the scoring function for each
  node $v \in N$, we can find a DAG whose moralized graph is a
  subgraph of $G$ maximizing the score in time and space $O(n)$.
\label{th1}
\end{myTheo}

We can combine the linear-time sampling of $k$-trees described in
Theorem~\ref{th0} with the linear-time learning of bounded structures
consistent with a graph in the above theorem to obtain an algorithm for
learning bounded treewidth Bayesian networks. The algorithm is described
in Algorithm~\ref{algo:method1} [Version 1].

\begin{algorithm}
\caption{Learning a structure of bounded treewidth by sampling
  Dandelion codes. There are two versions, according to the choice for
  step 2.c.}
\label{algo:method1}
\begin{algorithmic}
\item[\textbf{Input}]a score function $s_i, ~ \forall i\in N$.
\item[\textbf{Output}]a DAG $G^\text{best}$.
\item[1] Initialize $\pi_i^{\text{best}}$ as an empty set for all $i \in N$
\item[2] Repeat until a certain number of iterations is reached:
\item[2.a] Uniformly sample $(Q,S)\in\mathcal{A}_{n,k}$;
\item[2.b] Decode $(Q,S)$ into $T_k\in\mathcal{T}_{n,k}$;
\item[2.c] {\bf [Version 1]} Find a DAG $G$ that maximizes
  the score function and is {\it consistent} with $T_k$.
\item[2.c] {\bf [Version 2]} Sample $\sigma$ using
  Algorithm~\ref{method2}, and (greedily) find
  a DAG $G$ that maximizes the score function and is {\it consistent}
  with both $\sigma$ and $T_k$.
\item[2.d] If $\sum_{i\in N} s_i(\pi^G_i) > \sum_{i\in N}
  s_i(\pi_i^{\text{best}})$, update $\pi_i^{\text{best}}$, $\forall i$.
\end{algorithmic}
\end{algorithm}

\begin{myTheo}
The sampling space of Algorithm \ref{algo:method1} [Version 1] is less than
$e^{n\log (n k)}$. Each of its iterations runs in linear time in $n$ (but
exponential in $k$).
\end{myTheo}
\noindent {\it Proof.} The follow equality
holds~\cite{nktrees}.
\begin{equation}
|\mathcal{T}_{n,k}|=\binom{n}{k} \cdot \bigl(k(n-k)+1\bigr)^{n-k-2}.
\end{equation}
It is not hard to see that the maximum happens for $k\leq n/2$ (because
of the symmetry of $\binom{n}{k}$ and of $k(n-k)$ around $n/2$, while
$n-k-2$ decreases with the increase of $k$).  By manipulating this
number and applying Stirling's approximation for the factorials, we
obtain:
\begin{eqnarray}
|\mathcal{T}_{n,k}|&\!\!\!\!\leq \!\!\!\!& \frac{\sqrt{n} e^{n\log n+1
    -n}}{\left(\frac{n-k}{e}\right)^{n-k}\left(\frac{k}{e}\right)^k}
k^{n-k-2} (n-k)^{n-k-2}\nonumber\\
&\!\!\!\!\leq\!\!\!\!&\frac{e\sqrt{n}}{(n-k)^2}e^{n\log n} k^{n-2k-2}\leq e^{n\log n +
  (n-2k)\log k},\nonumber
\end{eqnarray}
which is less than $e^{n\log (n k)}$.  The decoding algorithm has
complexity linear in $n$ (Theorem~\ref{th0}), as well as the method to
uniformly sample a Dandelion code, and the method to find the best DAG
consistent with a $k$-tree (Theorem~\ref{th1}). $\blacksquare$

While the running time of Algorithm \ref{algo:method1} [Version 1] is
linear in $n$, the computational complexity of step 2.c, which uses the
method in~\cite{korhonen2exact}, is exponential in the treewidth (more
precisely, it is $\Theta(k\cdot 3^k\cdot (k+1)!\cdot n)$). Hence, one
cannot hope to use it with moderately high treewidth bounds (say, larger
than 8). Regarding the sample space, according to the above theorem it
is slightly higher than that of order-based learning of unconstrained
Bayesian networks (e.g.~\cite{teyssier2012ordering}), especially if
$k\ll n$. However, each iteration of step 2.c needs considerable more
effort than the corresponding iteration in the unbounded case (yet, as
it is a method theoretically linear in $n$, more efficient
implementations of the algorithm that searches within a given $k$-tree
might bring an additional boost to this approach in the future).

As just explained, the main practical drawback of Algorithm \ref{algo:method1} [Version 1]
is step 2.c, which process each sampled $k$-tree. In the sequel we
propose a new approach ([Version 2]) that is much faster (per
iteration), at the price of a slight increase in the sampling
space. We will empirically compare these approaches in the next section.

Let $\sigma$ define a partial order of the nodes. We say that a DAG $G$
is {\it consistent} with $\sigma$ if, $\forall j \in \text{pa}_i$ (as
defined by $G$), there is no directed path from $i$ to $j$ in
$\sigma$. In other words, $\sigma$ constrains the valid topological
orderings for the nodes in $G$.  We do not force $\sigma$ to be a linear
order, because we are only interested in orderings that specify, for
each edge in a $k$-tree $T_k$, which of the two ending points precedes
the other (in other words, we are only interested in possible ways of
orienting the edges of the k-tree). There are multiple linear orderings
that achieve the very same result for $T_k$, and our goal is to sample
from the smallest possible space of orderings (if we used a linear
order, then the sampling space would be $n!$).

A partial order $\sigma$ can be represented as a DAG $G$: $i$ is smaller
than $j$ in $\sigma$ if and only if node $i$ is an ancestor of node $j$
in $G$. Given a k-tree $T_k$, we will sample $\sigma$ by following the
same recursive process as in Definition~\ref{def1}. This is described in
Algorithm~\ref{method2}. The procedure produces partial orders (i.e.,
DAGs) whose underlying graph (obtained by ignoring arc directions) is
exactly the graph $T_k$. Note that the treewidth of the DAG
corresponding to $\sigma$ might exceed the treewidth of $k$. This does
not affect the correctness of Algorithm~\ref{algo:method1}, as $\sigma$
is only used to specify which node preceeds which node in the order, and
hence which are the possible parents; the actual parents are chosen so
that the treewidth bound is respect. This can be done efficiently using
$T_k$.

\begin{algorithm}
\caption{Sampling a partial order within a $k$-tree.}
\label{method2}
\begin{algorithmic}
\item[\textbf{Input}]a $k$-tree $T_k$ with $n$ nodes
\item[\textbf{Output}]a partial order defined by a DAG $\sigma$.
\item[1] Initialize $\sigma=T_k$. Arbitrarily choose a $(k+1)$-clique $R$ of
  $\sigma$ and call it the root clique;
\item[2] Uniformly sample the directions of the arcs in $\sigma$ linking the
  $k+1$ nodes in $R$, without creating cycles in $\sigma$, and mark these nodes as {\it done};
\item[3] Take a node $v$ that is linked to $k$ {\it done} nodes;
\item[4] Uniformly sample the directions of the arcs in $\sigma$ between $v$ and
  these $k$ {\it done} nodes, without creating cycles in $\sigma$, and mark $v$ as {\it done};
\item[5] Go to Step 3 unless all nodes are {\it done}.
\end{algorithmic}
\end{algorithm}

\begin{myTheo}
Algorithm \ref{method2} samples DAGs $\sigma$ on a sample
space of size $k!\cdot (k+1)^{n-k}$ and runs in linear time in
$n$ and $k$.
\end{myTheo}
\noindent {\it Proof.} The sampling of the $k+1$ nodes in the root
clique takes time $O(k)$ by sampling one of the $(k+1)!$ ways to choose
the arcs without creating cycles. We assume that an appropriate
structure representing $T_k$ is known (e.g., a tree-decomposition with
$n-k-1$ nodes), so Steps 1 and 3 can be done in $O(k)$ time.  For each
iteration of Step 4, we spend time $O(k)$ because there are only $k+1$
ways to direct the edges, as this is equivalent to placing $v$ in its
relative order with respect to the already ordered $k$ neighbors. Hence
the total running time is $O(k n)$ and the sampling space is
$(k+1)!\cdot (k+1)^{n-k-1}=k!\cdot (k+1)^{n-k}$. $\blacksquare$

The following result shows that the sampling space of this version of
the sampling algorithm remains reasonably small, especially for $k\ll n$
(it would be also small if $k$ is close to $n$, then $|\mathcal{T}_{n,k}|$ decreases
drastically, so the total sampling space would also decrease).
\begin{myTheo}
The sampling space of Algorithm \ref{algo:method1} [Version 2] is less than
$e^{n\log (n (k+1)^2)}$. Each of its iterations runs in linear
time in $n$ and $k$.
\end{myTheo}
\noindent {\it Proof.} As before, the decoding algorithm (Theorem~\ref{th0}) and
the method to uniformly sample a Dandelion code run in linear time in both $n$
and $k$. Algorithm~\ref{method2} samples the ordering $\sigma$ in linear time
too. Finally, finding the best DAG consistent with a $k$-tree $T_k$ and $\sigma$
is a greedy procedure over all nodes (choosing the parent set of a node each
time): the treewidth cannot exceed $k$ because we take a subgraph of $T_k$, and
no cycles can be formed if we respect $\sigma$. $\blacksquare$

Although the sampling space of Version 2 is larger than the one of
Version 1, Version 2 is much faster per iteration. This allows us to
explore a much larger region of the space of $k$-tress than Version 1 can
within a fixed amount of time. Moreover, one can run Version 2 without
pre-computing the score function: when scores are needed, they are
computed and stored into a hash table for further accesses, thus closely
matching another desirable characteristic of order-based learning
methods for unbounded treewidth (namely, to avoid computing all scores
a priori).

\section{Experiments}
\label{sec5}

\begin{table}
\centering
\caption{Dimensions of data sets.}
\begin{tabular}{lcc}
\toprule
DATA SET & VAR. & SAMPLES\\
\midrule
nursery &	9&12960\\
breast &	10&699\\
housing	&14&506\\
adult &	15&32561\\
zoo &	17&101\\
letter & 17&20000\\
mushroom & 22 & 8124\\
wdbc&31&569\\
audio &62&200\\
hill &100&606\\
community &100 & 1994\\
\bottomrule
\end{tabular}
\label{tablesize}
\end{table}
 
We empirically analyze the accuracy of Algorithm~\ref{algo:method1} by
comparing its two versions with each other and with the values obtained
by the MILP method. As before, we use a collection of data sets from the
UCI repository of varying dimensionality, with variables discretized
over the median value when needed.  The number of (binary) variables and
samples in each data set are described in Table~\ref{tablesize}. Some
columns of the original data sets {\it audio} and {\it community} were
discarded: 7 variables of {\it audio} had always a constant value, 5
variables of {\it community} have almost a different value per sample
(such as personal data), and 22 variables have missing data
(Table~\ref{tablesize} shows dimensions after this pre-processing). In
all experiments, we maximize the Bayesian Dirichlet likelihood
equivalent uniform (BDeu) score with equivalent sample size equal to
one~\cite{heckerman1995learning}.

We use treewidth bounds of 4 and 10, and maximum parent set size of 3
(for {\it hill} and {\it community}, it was set as 2;
nevertheless, the
MILP formulation is the one with a strong dependency on the maximum parent
set size, as scores need to be pre-computed). To be fair
among runs, we have pre-computed all scores, and have
considered them as input of the problem. The MILP has
been optimized by CPLEX 12.4 with a memory limit of 64GB. We have
allowed it to run up to three hours, and have also collected the
incumbent solution after 10 minutes. Algorithm~\ref{algo:method1} has
been given only 10 minutes (in either version).

\begin{figure}
  \centering

  \includegraphics{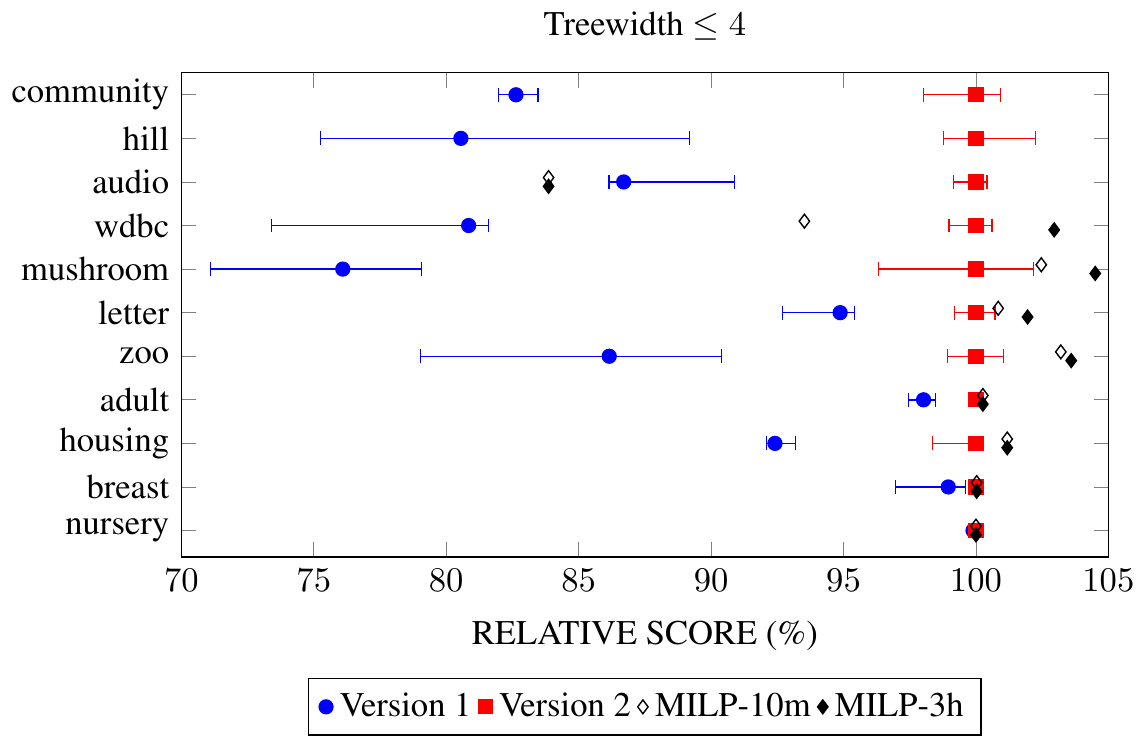}

  \caption{Performance of methods relative to the solution found by the
    Version 2 of Algorithm 1 with a treewidth limit of four. MILP
    results are missing for community and hill because it was not able
    to produce a solution for those cases.}
  \label{fig:tw4}
\end{figure}

\begin{figure}

  \centering

  \includegraphics[width=11cm]{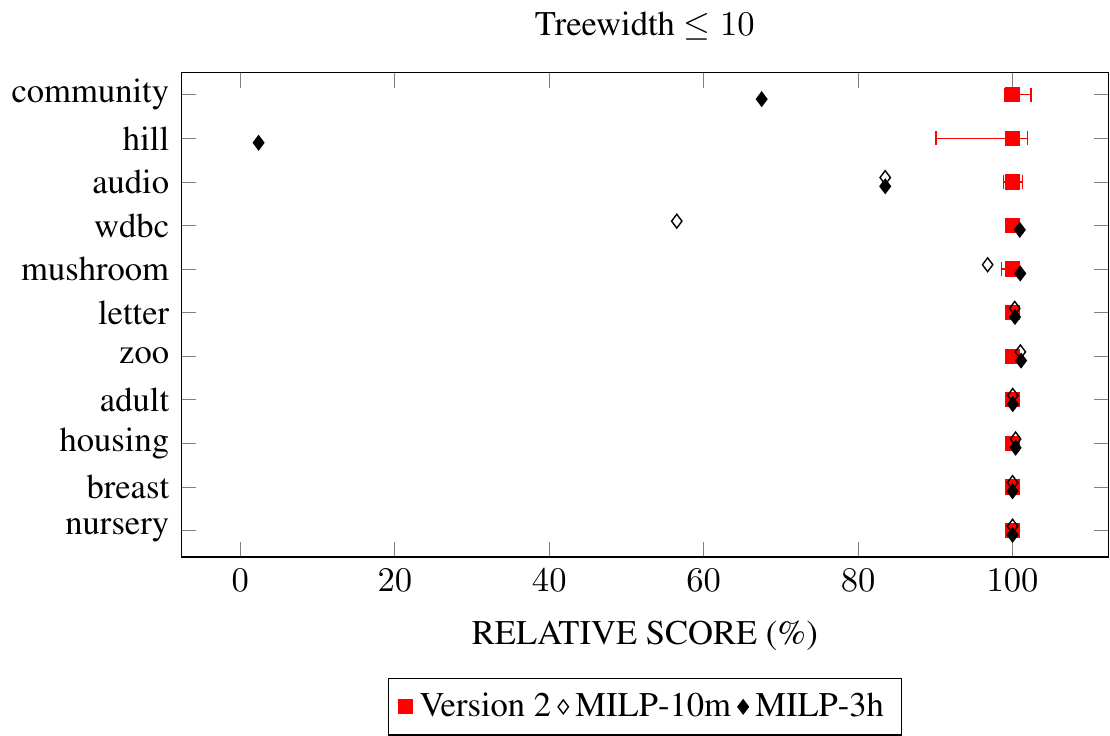}

  \caption{Performance of methods relative to the solution found by the
    Version 2 of Algorithm 1 with a treewidth limit of ten. MILP
    results after 10 minutes
    are missing for community and hill because it was not able to
    produce a solution within that time.}
  \label{fig:tw10}
\end{figure}

To account for the variability of the performance of the sampling
methods with respect to the sampling seed, we ran each version of
Algorithm~\ref{algo:method1} ten times on each data set with different
seeds. We report the minimum, median and maximum obtained values over
those runs for each dataset.  We show the relative scores (in
percentage) of the approximate methods (Versions 1 and 2 of
Algorithm~\ref{algo:method1} and the best score found by the MILP
formulation within 10 minutes and 3 hours) with respect to Version 2's
median score, for treewidth bounds of four (Figure~\ref{fig:tw4}) and
ten (Figure~\ref{fig:tw10}). The relative score is computed as the ratio
of the obtained value and the median score of Version 2, so higher
values are better. Moreover, a value higher than 100\% shows that the
method outperformed Version 2, whereas a value smaller than 100\% shows
the converse. The raw data used in the figures appear in
Tables~\ref{table1a} (for Figure~\ref{fig:tw4}) and~\ref{table2a} (for
Figure~\ref{fig:tw10}). The exponential dependence on treewidth of
Version 1 made it intractable to run with treewidth bound greater than
8.  We see from the plot on top that Version 2 is largely superior to
Version 1, even if the former might only find suboptimal networks for a
given $k$-tree. This is probably a consequence of the much lower running
times per iteration, which allows Version 2 to explore a much larger set
of $k$-trees. It also suggests that spending time finding good $k$-trees
is more worthy than optimizing network structures for a given
$k$-tree. We also see that the MILP formulation scales poorly with the
number of variables, being unable to obtain satisfactory solutions for
data sets with more than 50 variables. On the {\it hill} data set with
treewidth $\leq 4$, CPLEX running the MILP formulation was not able to
output any solution within 10 minutes, and the solution obtained within
3 hours is far left of the zoomed area of the graph in
Figure~\ref{fig:tw4}; on the {\it community} data set with treewidth
$\leq 4$, CPLEX did not find any solution within 3 hours. Regarding the
treewidth bound of ten (Figure~\ref{fig:tw10}), we observe that Version
2 is very accurate and outperforms the MILP formulation in the larger
data sets.

It is worth noting that both versions of Algorithm~\ref{algo:method1}
were implemented in Matlab; hence, the comparison with the approximate
solution of running the MILP formulation with the same amount of time
(10 minutes) might be unfair, as we expect to produce better results by
an appropriate re-coding of our sampling methods in a more efficient
language (one could also try to improve the MILP formulation, although
it will eventually suffer from the problems discussed in
Section~\ref{sec3}).  Nevertheless, the results show that Version 2 is
very competitive even in this scenario.

\section{Conclusions}
\label{sec6}

We have created new exact and approximate procedures to learn Bayesian
networks of bounded treewidth. They perform well and are of immediate
practical use. The designed mixed-integer linear programming (MILP)
formulation improves on MILP formulations for related tasks, especially
regarding the specification of treewidth-related constraints. It solves
the problem exactly and surpasses a state-of-the-art method both in size
of networks and treewidth that it can handle. Even if results indicate
it is better than the state of the art, MILP is not so accurate and
might fail in large domains. For that purpose, we have proposed a double
sampling idea that provides means to learn Bayesian networks in large
domains and high treewidth limits, and is empirically shown to perform
very well in a collection of public data sets. It scales well, because
its complexity is linear both in the domain size and in the treewidth
bound. There are certainly other search methods that can be integrated
with our sampling approach, for instance a local search after every
iteration of sampling, local permutations of orderings that are
compatible with the $k$-trees, etc. We leave the study of these and
other avenues for future work.

During the making of this work, two closely related works appeared in
the literature. \cite{bjm2014aistats} developed an exact learning
procedure based on maximum satisfiability. \cite{pfl2014aistats}
developed an alternative MILP formulation of the problem with
exponentially many constraints, and used cutting plane generation
techniques to improve on performance. These works have been developed
independently and simultaneously with our work presented here; future
work should compare their performance empirically against the methods
proposed here.

\section{Acknowledgments}

This work was partly supported by the grant N00014-12-1-0868 from the
US Office of Navy Research, the Swiss NSF grant n. 200021\_146606/1, and
the FAPESP grant n. 2013/23197-4.

\bibliography{refs}
\bibliographystyle{plainnat}

\begin{sidewaystable}
\centering
\caption{Performance of methods within 10 minutes of
time limit (except for last column, which is 3 hours) and treewidth
limit of four. Maximum, median and minimum are respect to 10 runs with
different random seeds. MILP results are missing when it was not
able to produce a solution within the time limit. When the incumbent solution is
not yet proven optimal, after the last
column there is the maximum error of the incumbent solution.}
\bigskip
\begin{tabular}{c|ccc|ccc|cc}
\toprule
 & \multicolumn{3}{c|}{Algorithm 1 -- Version 1} &
 \multicolumn{3}{c|}{Algorithm 1 -- Version 2}& \multicolumn{2}{c}{MILP}\\
DATSET & MAX & MIN & MEDIAN & MAX &  MIN & MEDIAN & 10m & 3h\\
\midrule
    nursery & -72235.68 & -72327.59 & -72239.34 & -72159.27 & -72159.27 & -72159.27 & -72159.27 & -72159.27\\
    breast & -2696.62 & -2770.36 & -2714.39 & -2685.24 & -2686.87 & -2685.99 & -2685.24 & -2685.24\\
    housing & -3430.32 & -3470.83 & -3458.86 & -3186.99 & -3249.59 & -3196.41 & -3159.1 & -3159.1\\
    adult & -204063.81 & -206209.91 & -205008.65 & -200771.07 & -201277.83 & -200950.56 & -200426.42 & -200426.42\\
    zoo   & -644.55 & -737.13953 & -676.20 & -576.58 & -588.89 & -582.57 & -564.49 & -562.35\\
    letter & -196931.49 & -202689.46 & -198054.51 & -186553.59 & -189419.55 & -187901.06 & -186335.69 & -184308.01\\
    mushroom & -76431.11 & -84982.26 & -79416.18 & -59147.71 &
    -62745.76 & -60431.06 & -58976.54 & -57827.36 [14.9\%]\\
    wdbc & -8825.27 & -9810.43 & -8907.39 & -7158.26 & -7274.97 &
    -7201.37 & -7700.21 & -6994.85 [9.0\%]\\
    audio & -2345.23 & -2473.97 & -2458.18 & -2122.33 & -2149.55 &
    -2131.19 & -2541.25 & -2541.25 [23.5\%]\\
    hill    & -1275.01 & -1510.71 & -1411.53& -1112.05 &   -1151.12 &
    -1137.03 & -- & -42347.68 [99.1\%]\\
    community & -111792.03 & -113826.56 & -112910.15 & -92433.71 & -95183.13 & -93300.99 & -- & --\\
\bottomrule
\end{tabular}
\label{table1a}
\end{sidewaystable}

\begin{sidewaystable}
\centering
\caption{Performance of methods within 10 minutes of
time limit (except for last column, which is 3 hours) and treewidth
limit of ten. Maximum, median and minimum are respect to 10 runs with
different random seeds. MILP results are missing when it was not
able to produce a solution within the time limit. When the incumbent solution is
not yet proven optimal, after the last
column there is the maximum error of the incumbent solution.}
\bigskip
\begin{tabular}{c|ccc|cc}
\toprule
 & \multicolumn{3}{c|}{Algorithm 1 -- Version 2}& \multicolumn{2}{c}{MILP}\\
DATASET & MAX & MIN & MEDIAN & 10m & 3h\\
\midrule
    nursery & -72159.27 & -72159.27 & -72159.27 & -72159.27 & -72159.27\\
    breast & -2685.24 & -2685.24 & -2685.24 & -2685.24 & -2685.24\\
    housing & -3159.71 & -3180.67 & -3171.55 & -3159.1 & -3159.1\\
    adult & -200398.91 & -200513.6 & -200419.75 & -200362.49 & -200362.49\\
    zoo & -564.24 & -572.13 & -567.76 &-562.08  & -561.54\\
    letter & -184077.80 & -184840.36 & -184552.17 & -184030.33 & -183972.77 \\
    mushroom & -55604.24 & -56950.21 & -56116.23 & -57987.74 &
    -55562.44 [9.8\%]\\
    wdbc & -6943.96 & -6984.57 & -6958.57 & -12313.03 & -6893.55 [7.5\%]\\
    audio & -2094.76 & -2146.76 & -2122.17 & -2541.25 & -2541.25 [23.8\%]\\
    hill & -980.06 & -1109.28 & -999.11 & -- & -42347.68 [99.1\%]\\
    community & -89556.12 & -92656.88 & -91709.72 & -- & -135844.68 [45.9\%]\\
\bottomrule
\end{tabular}
\label{table2a}
\end{sidewaystable}

\end{document}